\documentclass[11pt,a4paper]{article}
\usepackage[nohyperref]{naaclhlt2018}
\usepackage{times}
\usepackage{latexsym}

\usepackage{graphicx}
\usepackage[table,xcdraw]{xcolor}
\usepackage{booktabs}
\usepackage{multirow}
\usepackage{url}
\usepackage{svg}

\usepackage{color, soul}

\definecolor{purplex}{HTML}{dadaeb}
\definecolor{greenx}{HTML}{78c679}
\definecolor{bvd-green}{HTML}{B7FEA6}
\definecolor{bvd-blue}{HTML}{A6D1FE}
\definecolor{bvd-gray}{HTML}{EEE9FD}
\definecolor{orangex}{HTML}{fee090}

\newcommand{\hlc}[2][yellow]{ {\sethlcolor{#1} \hl{#2}} }

\newcommand\capsize{\capsizeinner}

\newlength\defaultparindent
\aclfinalcopy 
\def\aclpaperid{1017} 

\title{Gender Bias in Coreference Resolution}

\author{Rachel Rudinger,\hspace{.3em} Jason Naradowsky,\hspace{.3em} Brian Leonard,\hspace{.3em} and Benjamin Van Durme \\
  Johns Hopkins University\\}

\date{}

\begin{document}
\maketitle
\begin{abstract}
We present an empirical study of gender bias in coreference resolution systems. We first introduce a novel, Winograd schema-style set of minimal pair sentences that differ only by pronoun gender.
With these \textit{Winogender schemas}, we evaluate and confirm systematic gender bias in three publicly-available coreference resolution systems, and correlate this bias with real-world and textual gender statistics.
\end{abstract}

\section{Introduction}
\label{sec:intro}

There is a classic riddle: \textit{A man and his son get into a terrible car crash. The father dies, and the boy is badly injured. In the hospital, the surgeon looks at the patient and exclaims, ``I can't operate on this boy, he's my son!''} \textbf{How can this be?}

That a majority of people are reportedly unable to solve this riddle\footnote{The surgeon is the boy's mother.} is taken as evidence of underlying implicit gender bias \cite{wapman}:  many first-time listeners have difficulty assigning both the role of ``mother'' and ``surgeon'' to the same entity.

\begin{figure}
  \centering
    \fbox{\includegraphics[width=0.47\textwidth]{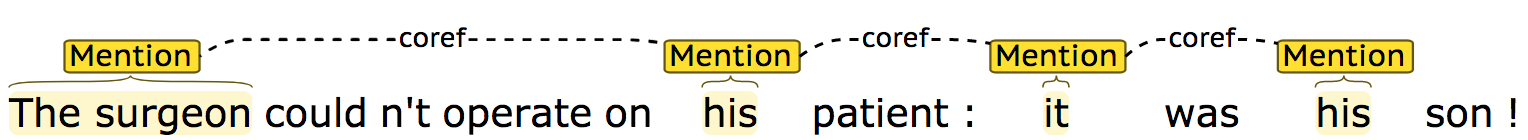}}
    \fbox{\includegraphics[width=0.47\textwidth]{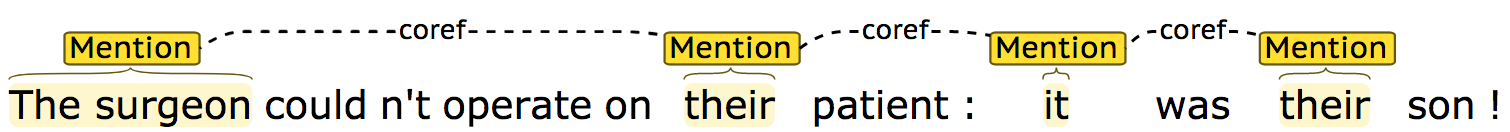}}
    \fbox{\includegraphics[width=0.47\textwidth]{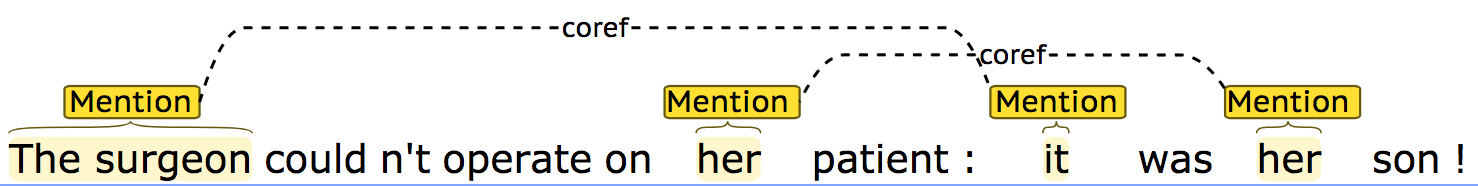}}
  \caption{\capsize Stanford CoreNLP rule-based coreference system resolves a male and neutral pronoun as coreferent with ``The surgeon,'' but does not for the corresponding female pronoun.}
  \label{fig:surgeon}
\end{figure}

As the riddle reveals, the task of coreference resolution in English is tightly bound with questions of gender, for humans and automated systems alike (see Figure \ref{fig:surgeon}).
As awareness grows of the ways in which data-driven AI technologies may acquire and amplify human-like biases \cite{Caliskan183,barocas2016big,hovy-spruit:2016:P16-2}, this work investigates how gender biases manifest in coreference resolution systems.

There are many ways one could approach this question; here we focus on gender bias with respect to occupations, for which we have corresponding U.S. employment statistics.
Our approach is to construct a challenge dataset in the style of \emph{Winograd schemas}, wherein a pronoun must be resolved to one of two previously-mentioned entities in a sentence designed to be easy for humans to interpret, but challenging for data-driven systems~\cite{Levesque2011TheWS}. In our setting, one of these mentions is a person referred to by their occupation; by varying only the pronoun's gender, we are able to test the impact of gender on resolution. With these ``Winogender schemas,'' we demonstrate the presence of systematic gender bias in multiple publicly-available coreference resolution systems, and that occupation-specific bias is correlated with employment statistics. We release these test sentences to the public.\footnote{\url{https://github.com/rudinger/winogender-schemas}}

In our experiments, we represent gender as a categorical variable with either two or three possible values: female, male, and (in some cases) neutral. These choices reflect limitations of the textual and real-world datasets we use.

\section{Coreference Systems}
\label{sec:background-systems}

In this work, we evaluate three publicly-available off-the-shelf coreference resolution systems, representing three different machine learning paradigms: rule-based systems, feature-driven statistical systems, and neural systems.

\paragraph{Rule-based}
In the absence of large-scale data for training coreference models, early systems relied heavily on expert knowledge.  A frequently used example of this is the Stanford multi-pass sieve system~\citep{lee11conllst}.  A deterministic system, the sieve consists of multiple rule-based models which are applied in succession, from highest-precision to lowest.  Gender is among the set of mention attributes identified in the very first stage of the sieve, making this information available throughout the system. 

\paragraph{Statistical}
Statistical methods, often with millions of parameters, ultimately surpassed the performance of rule-based systems on shared task data~\citep{DurrettKlein2013, bjorkelund-kuhn:2014:P14-1}.  The system of \citet{DurrettKlein2013} replaced hand-written rules with simple feature templates.  Combinations of these features implicitly capture linguistic phenomena useful for resolving antecedents, but they may also unintentionally capture bias in the data.  For instance, for occupations which are not frequently found in the data, an occupation+pronoun feature can be highly informative, and the overly confident model can exhibit strong bias when applied to a new domain.

\paragraph{Neural}

The move to deep neural models led to more powerful antecedent scoring functions, and the subsequent learned feature combinations resulted in new state-of-the-art performance~\citep{P15-1137, clark2016improving}.  Global inference over these models further improved performance~\citep{wiseman-rush-shieber:2016:N16-1, clark2016deep}, but from the perspective of potential bias, the information available to the model is largely the same as in the statistical models. A notable exception is in the case of systems which make use of pre-trained word embeddings~\citep{clark2016improving}, which have been shown to contain bias and have the potential to introduce bias into the system.

\paragraph{Noun Gender and Number}
Many coreference resolution systems, including those described here, make use of a common resource released by \newcite{bergsma-lin:2006:COLACL}\footnote{This data was distributed in the CoNLL 2011 and 2012 shared tasks on coreference resolution. \cite{pradhan-etal-conll-st-2011-ontonotes,pradhan-etal-conll-st-2012-ontonotes}} (``B\&L''): a large list of English nouns and noun phrases with gender and number counts over 85GB of web news. For example, according to the resource, 9.2\% of mentions of the noun ``doctor'' are female. The resource was compiled by bootstrapping coreference information from the dependency paths between pairs of pronouns. We employ this data in our analysis.

\section{Winogender Schemas}
\label{sec:data}

Our intent is to reveal cases where coreference systems may be more or less likely to recognize a pronoun as coreferent with a particular occupation based on pronoun gender, as observed in Figure \ref{fig:surgeon}.
To this end, we create a specialized evaluation set consisting of 120 hand-written sentence templates, in the style of the Winograd Schemas~\cite{Levesque2011TheWS}. Each sentence contains three referring expressions of interest:
\begin{enumerate}
\item \hlc[bvd-green]{\textbf{\textsc{occupation}}}, a person referred to by their occupation and a definite article, e.g., ``the paramedic.''
\item \hlc[bvd-blue]{\textbf{\textsc{participant}}}, a secondary (human) participant, e.g., ``the passenger.''
\item \hlc[bvd-gray]{\textbf{\textsc{pronoun}}}, a pronoun that is coreferent with either \textsc{occupation} or \textsc{participant}.
\end{enumerate}

We use a list of 60 one-word occupations obtained from \newcite{Caliskan183} (see supplement), with corresponding gender percentages available from the U.S. Bureau of Labor Statistics.\footnote{50 are from the supplement of \newcite{Caliskan183}, an additional 7 from personal communication with the authors, and three that we added: \textit{doctor}, \textit{firefighter}, and \textit{secretary}.} For each occupation, we wrote two similar sentence templates: one in which \textsc{pronoun} is coreferent with \textsc{occupation}, and one in which it is coreferent with \textsc{participant} (see Figure \ref{fig:sentencepair}).
For each sentence template, there are three \textsc{pronoun} instantiations (female, male, or neutral), and two \textsc{participant} instantiations (a specific participant, e.g., ``the passenger,'' and a generic paricipant, ``someone.'') With the templates fully instantiated, the evaluation set contains 720 sentences: 60 occupations $\times$ 2 sentence templates per occupation $\times$ 2 participants $\times$ 3 pronoun genders.

\begin{figure}[t]
{
\small
\noindent       
\fbox{%
    \parbox{.97\linewidth\fboxsep-1.6pt}{%
    \parindent\defaultparindent%

\indent (1a) \hlc[bvd-green]{\textbf{The paramedic}} performed CPR on \hlc[bvd-blue]{the passenger} even though \hlc[bvd-gray]{she/he/they} knew it was too late.\vspace{1mm}

(2a) \hlc[bvd-green]{The paramedic} performed CPR on \hlc[bvd-blue]{\textbf{the passenger}} even though \hlc[bvd-gray]{she/he/they} was/were already dead.\vspace{1mm}

(1b) \hlc[bvd-green]{\textbf{The paramedic}} performed CPR on \hlc[bvd-blue]{someone} even though \hlc[bvd-gray]{she/he/they} knew it was too late.\vspace{1mm}

(2b) \hlc[bvd-green]{The paramedic} performed CPR on \hlc[bvd-blue]{\textbf{someone}} even though \hlc[bvd-gray]{she/he/they} was/were already dead.
}}
}
\caption{\capsize A ``Winogender'' schema for the occupation \textit{paramedic}. Correct answers in bold. In general, \textsc{occupation} and \textsc{participant} may appear in either order in the sentence.}
\label{fig:sentencepair}
\end{figure}

\paragraph{Validation}

Like Winograd schemas, each sentence template is written with one intended correct answer (here, either \textsc{occupation} or \textsc{participant}).\footnote{Unlike Winograd schemas, we are not primarily concerned with whether these sentences are ``hard'' to solve, e.g., because they would require certain types of human knowledge or could not be easily solved with word co-occurrence statistics.}
We aimed to write sentences where (1) pronoun resolution was as unambiguous for humans as possible (in the absence of additional context), and (2) the resolution would not be affected by changing pronoun gender. (See Figure \ref{fig:sentencepair}.)
Nonetheless, to ensure that our own judgments are shared by other English speakers, we validated all 720 sentences on Mechanical Turk, with 10-way redundancy.
Each MTurk task included 5 sentences from our dataset, and 5 sentences from the Winograd Schema Challenge \cite{Levesque2011TheWS}\footnote{We used the publicly-available examples from \url{https://cs.nyu.edu/faculty/davise/papers/WinogradSchemas/WSCollection.html}}, though this additional validation step turned out to be unnecessary.\footnote{In the end, we did not use the Winograd schemas to filter annotators, as raw agreement on the Winogender schemas was much higher to begin with (94.9\% Winogender vs. 86.5\% Winograd).} 
Out of 7200 binary-choice worker annotations (720 sentences $\times$ 10-way redundancy), 94.9\% of responses agree with our intended answers.
With simple majority voting on each sentence, worker responses agree with our intended answers for 718 of 720 sentences (99.7\%). The two sentences with low agreement have neutral gender (``they''), and are not reflected in any binary (female-male) analysis.

\begin{figure}[tb]
  \centering
  \includegraphics[width=.45\textwidth]{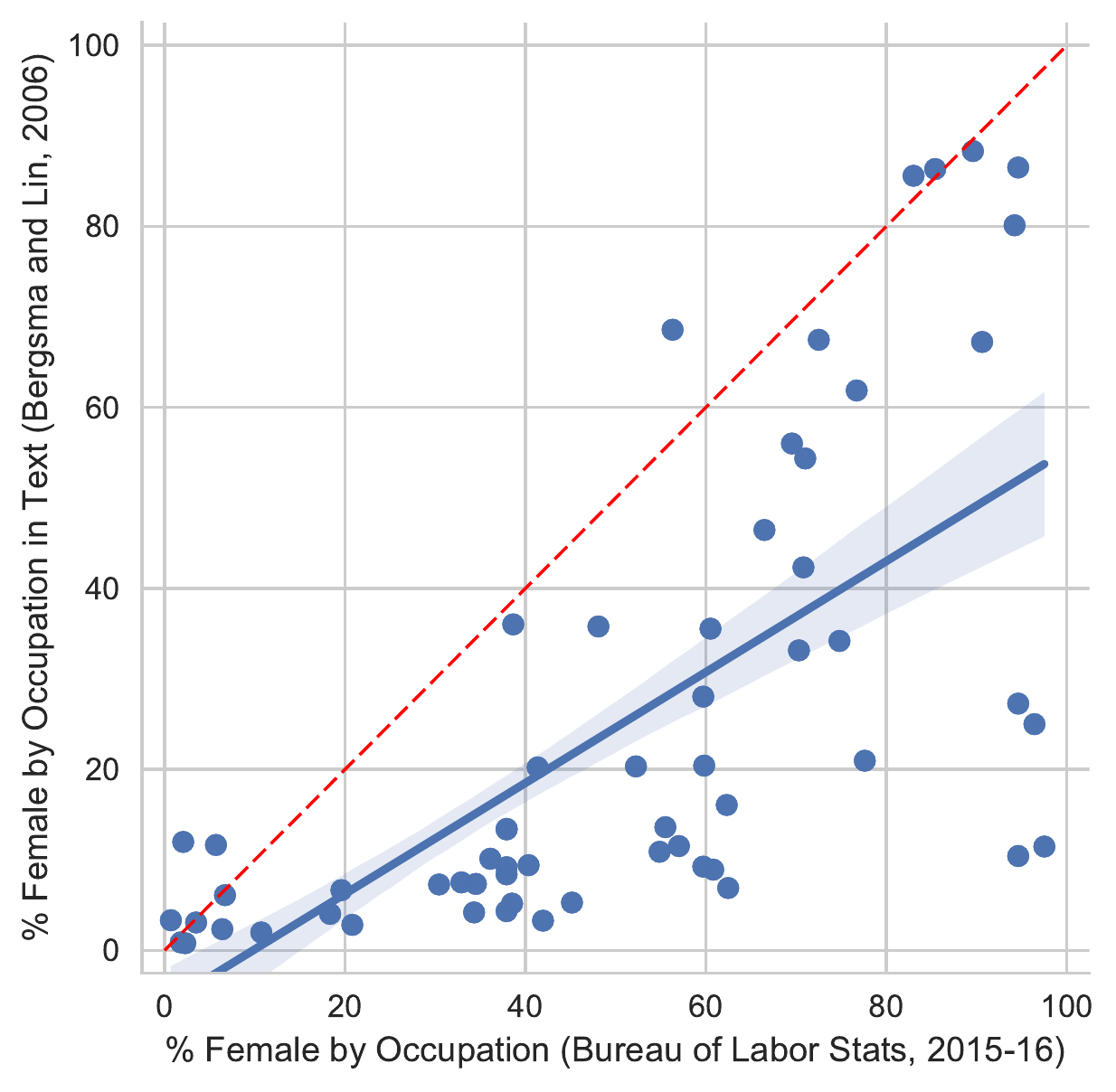}
  \caption{\capsize Gender statistics from \newcite{bergsma-lin:2006:COLACL} correlate with Bureau of Labor Statistics 2015. However, the former has systematically lower female percentages; most points lie well below the 45-degree line (dotted). Regression line and 95\% confidence interval in blue. Pearson r = 0.67.}
\label{fig:bergsmabls}
\end{figure}

\begin{table}[htb]
\begin{center}
\begin{tabular}{lccc}
\toprule
Correlation (r) & \textsc{Rule} & \textsc{Stat} & \textsc{Neural}  \\
\midrule
B\&L &                0.87 &    0.46 & 0.35\\
BLS &                0.55 &     0.31 & 0.31\\
\bottomrule
\end{tabular}
\end{center}
\caption{\capsize Correlation values for Figures \ref{fig:bergsmabls} and \ref{fig:scatter}.}
\label{tab:corr}
\end{table}

\section{Results and Discussion}

\begin{figure*}[htb]
\centering
\begin{minipage}{.45\textwidth}
  \centering
  \includegraphics[width=\textwidth]{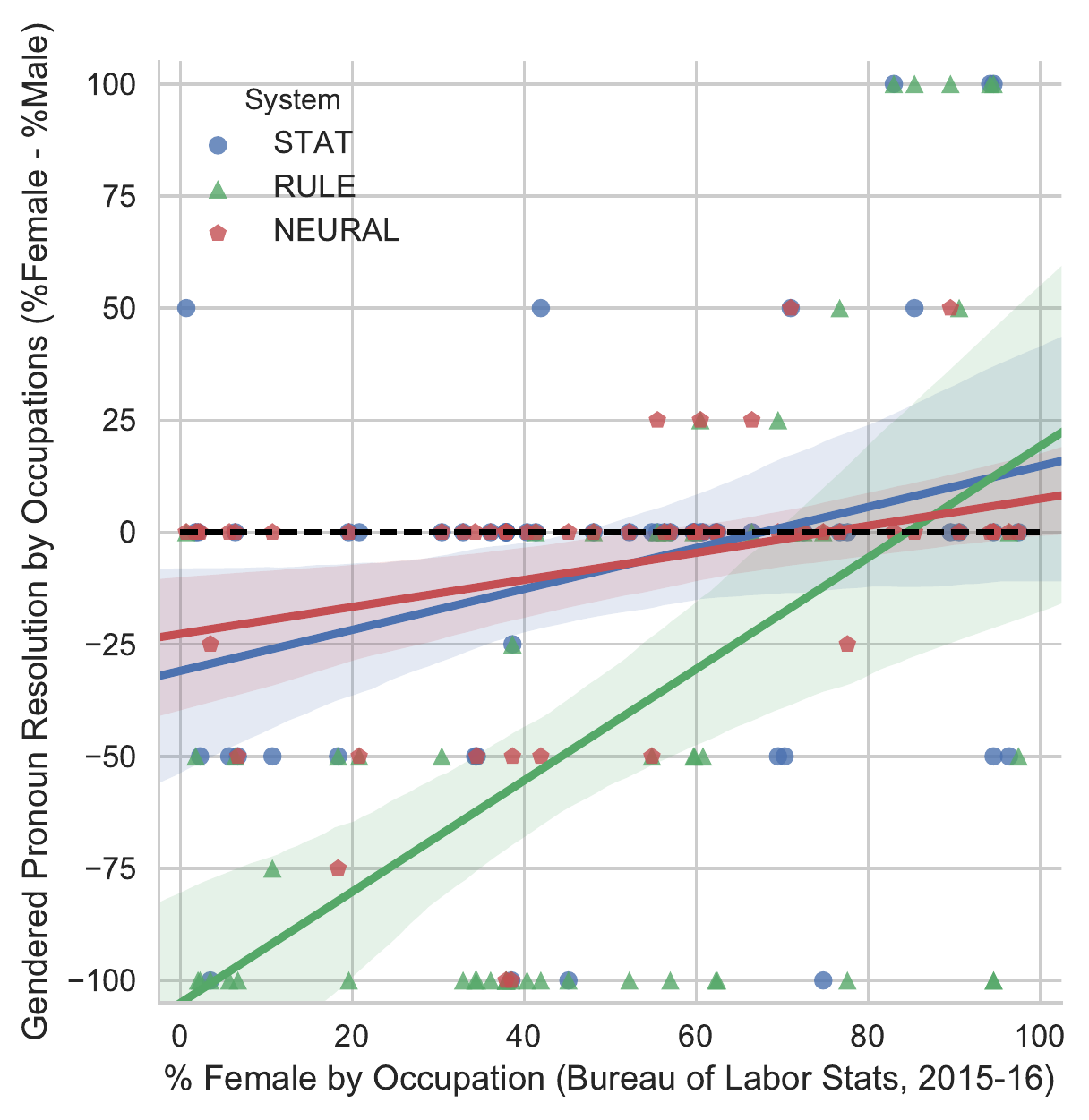}
\end{minipage}
\hspace{2em}
\begin{minipage}{.45\textwidth}
  \centering
  \includegraphics[width=\textwidth]{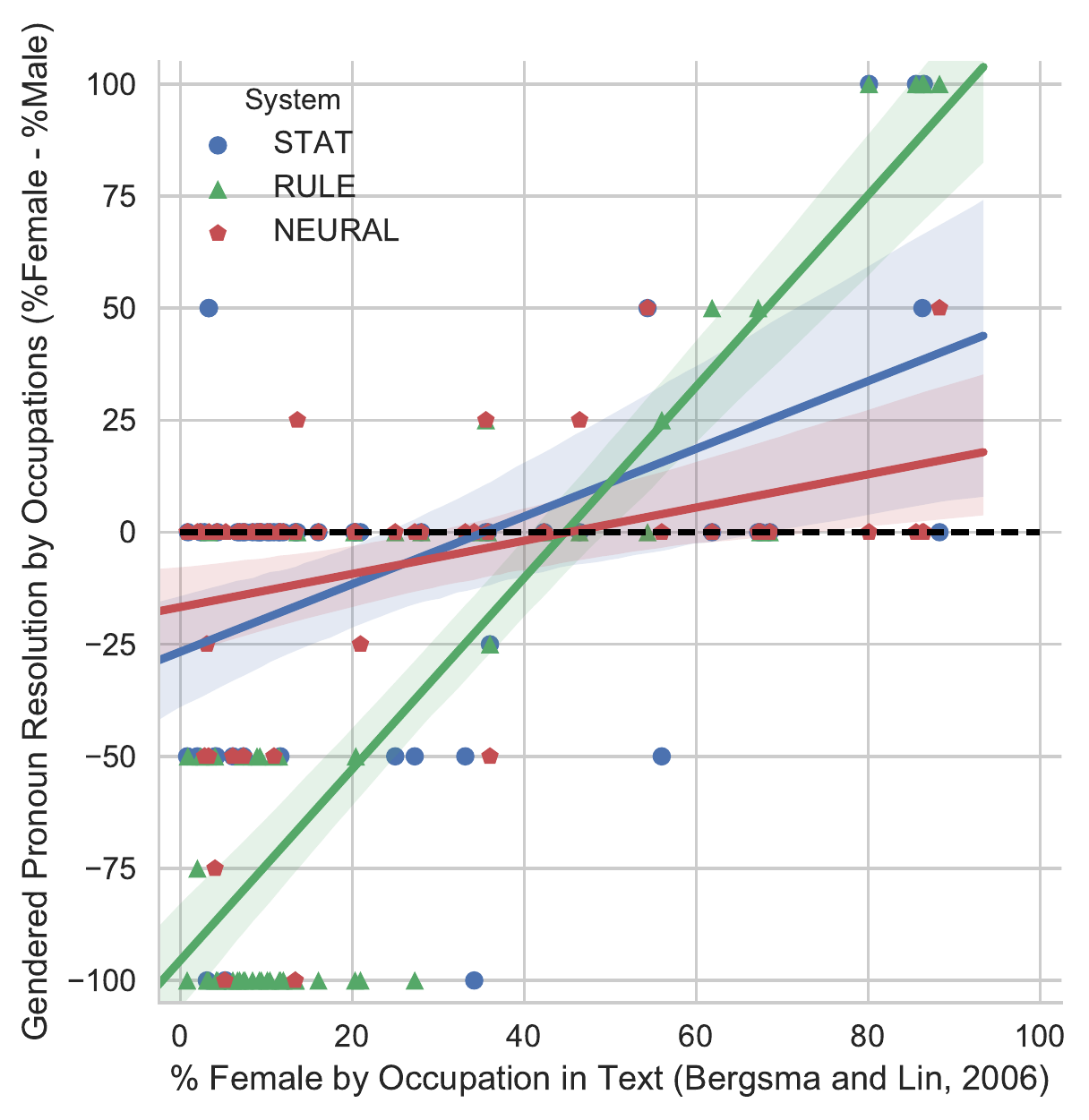}
\end{minipage}
\caption{\capsize These two plots show how gender bias in coreference systems corresponds with occupational gender statistics from the U.S Bureau of Labor Statistics (left) and from text as computed by \newcite{bergsma-lin:2006:COLACL} (right); each point represents one occupation. The y-axes measure the extent to which a coref system prefers to match female pronouns with a given occupation over male pronouns, as tested by our Winogender schemas. A value of 100 (maximum female bias) means the system always resolved female pronouns to the given occupation and never male pronouns (100\% - 0\%); a score of -100 (maximum male bias) is the reverse; and a value of 0 indicates no gender differential. Recall the Winogender evaluation set is gender-balanced for each occupation; thus the horizontal dotted black line (y=0) in both plots represents a hypothetical system with 100\% accuracy. Regression lines with 95\% confidence intervals are shown.}
\label{fig:scatter}
\end{figure*}

We evaluate examples of each of the three coreference system architectures described in \ref{sec:background-systems}: the \citet{lee11conllst} sieve system from the rule-based paradigm (referred to as {\sc RULE}), \citet{DurrettKlein2013} from the statistical paradigm ({\sc STAT}), and the \citet{clark2016deep} deep reinforcement system from the neural paradigm ({\sc NEURAL}).

By multiple measures, the Winogender schemas reveal varying degrees of gender bias in all three systems. First we observe that these systems do not behave in a gender-neutral fashion. That is to say, we have designed test sentences where correct pronoun resolution is not a function of gender (as validated by human annotators), but system predictions do exhibit sensitivity to pronoun gender: 68\% of male-female minimal pair test sentences are resolved differently by the \textsc{RULE} system; 28\% for \textsc{STAT}; and 13\% for \textsc{NEURAL}.

Overall, male pronouns are also more likely to be resolved as \textsc{occupation} than female or neutral pronouns across all systems: for \textsc{RULE}, 72\% male vs 29\% female and 1\% neutral; for \textsc{STAT}, 71\% male vs 63\% female and 50\% neutral; and for \textsc{NEURAL}, 87\% male vs 80\% female and 36\% neutral. Neutral pronouns are often resolved as neither \textsc{occupation} nor \textsc{participant}, possibly due to the number ambiguity of ``they/their/them.''

When these systems' predictions diverge based on pronoun gender, they do so in ways that reinforce and magnify real-world occupational gender disparities. Figure \ref{fig:scatter} shows that systems' gender preferences for occupations correlate with real-world employment statistics (U.S. Bureau of Labor Statistics) and the gender statistics from text \cite{bergsma-lin:2006:COLACL} which these systems access directly; correlation values are in Table \ref{tab:corr}. We also identify so-called ``gotcha'' sentences in which pronoun gender does not match the occupation's majority gender (BLS) if \textsc{occupation} is the correct answer; all systems perform worse on these ``gotchas.''\footnote{``\hspace{-0.3em}\hlc[bvd-green]{\textbf{The librarian}} helped \hspace{-0.3em}\hlc[bvd-blue]{the child} pick out a book because \hlc[bvd-gray]{he} liked to encourage reading.'' is an example of a ``gotcha'' sentence; librarians are $>50\%$ female (BLS).} (See Table \ref{tab:hard}.)

Because coreference systems need to make discrete choices about which mentions are coreferent, percentage-wise differences in real-world statistics may translate into absolute differences in system predictions. For example, the occupation ``manager'' is 38.5\% female in the U.S. according to real-world statistics (BLS); mentions of ``manager'' in text are only 5.18\% female (B\&L resource); and finally, as viewed through the behavior of the three coreference systems we tested, no managers are predicted to be female.
This illustrates two related phenomena: first, that data-driven NLP pipelines are susceptible to sequential amplification of bias throughout a pipeline, and second, that although the gender statistics from B\&L correlate with BLS employment statistics, they are systematically male-skewed (Figure \ref{fig:bergsmabls}).

\begin{table}[htb]
\begin{center}
\small
\begin{tabular}{lccc}
\toprule
System & ``Gotcha''? &   Female      &   Male    \\
\midrule
\multicolumn{1}{c}{\multirow{2}{*}{\textsc{Rule}}} & no &    38.3 &  51.7 \\
                          & \cellcolor[HTML]{CBCEFB}yes  &    \cellcolor[HTML]{CBCEFB}10.0 &  \cellcolor[HTML]{CBCEFB}37.5 \\
\hline
\multicolumn{1}{c}{\multirow{2}{*}{\textsc{Stat}}} & no &    50.8 &  61.7 \\
                          & \cellcolor[HTML]{CBCEFB}yes  &    \cellcolor[HTML]{CBCEFB}45.8 &  \cellcolor[HTML]{CBCEFB}40.0 \\
\hline

\multicolumn{1}{c}{\multirow{2}{*}{\textsc{Neural}}} & no &    50.8 &  49.2 \\
                          & \cellcolor[HTML]{CBCEFB}yes  &    \cellcolor[HTML]{CBCEFB}36.7 &  \cellcolor[HTML]{CBCEFB}46.7 \\
\bottomrule
\end{tabular}
\end{center}
\caption{\capsize System accuracy (\%) bucketed by gender and difficulty (so-called ``gotchas,'' shaded in purple). For female pronouns, a ``gotcha'' sentence is one where either (1) the correct answer is \textsc{occupation} but the occupation is $<50\%$ female (according to BLS); or (2) the occupation is $\geq 50\%$ female but the correct answer is \textsc{participant}; this is reversed for male pronouns. Systems do uniformly worse on ``gotchas.''}
\label{tab:hard}
\end{table}
\vspace{-1.4em}

\section{Related Work}

Here we give a brief (and non-exhaustive) overview of prior work on gender bias in NLP systems and datasets. A number of papers explore (gender) bias in English word embeddings: how they capture implicit human biases in modern \cite{Caliskan183} and historical \cite{Garg201720347} text, and methods for debiasing them \cite{NIPS2016_6228}. Further work on debiasing models with adversarial learning is explored by \newcite{DBLP:journals/corr/BeutelCZC17} and \newcite{zhang2018mitigating}.

Prior work also analyzes social and gender stereotyping in existing NLP and vision datasets \cite{van2016stereotyping,rudinger-may-vandurme:2017:EthNLP}. \newcite{tatman:2017:EthNLP} investigates the impact of gender and dialect on deployed speech recognition systems, while \newcite{zhao-EtAl:2017:EMNLP20173} introduce a method to reduce amplification effects on models trained with gender-biased datasets. \newcite{koolen-vancranenburgh:2017:EthNLP} examine the relationship between author gender and text attributes, noting the potential for researcher interpretation bias in such studies.
Both \newcite{larson:2017:EthNLP} and \newcite{koolen-vancranenburgh:2017:EthNLP} offer guidelines to NLP researchers and computational social scientists who wish to predict gender as a variable. \newcite{hovy-spruit:2016:P16-2} introduce a helpful set of terminology for identifying and categorizing types of bias that manifest in AI systems, including \textit{overgeneralization}, which we observe in our work here.

Finally, we note independent but closely related work by \newcite{zhao-wang:2018:N18-1}, published concurrently with this paper. In their work, \newcite{zhao-wang:2018:N18-1} also propose a Winograd schema-like test for gender bias in coreference resolution systems (called ``WinoBias''). Though similar in appearance, these two efforts have notable differences in substance and emphasis. The contribution of this work is focused primarily on schema construction and validation, with extensive analysis of  observed system bias, revealing its correlation with biases present in real-world and textual statistics; by contrast, \newcite{zhao-wang:2018:N18-1} present methods of debiasing existing systems, showing that simple approaches such as augmenting training data with gender-swapped examples or directly editing noun phrase counts in the B\&L resource are effective at reducing system bias, as measured by the schemas. Complementary differences exist between the two schema formulations: Winogender schemas (this work) include gender-neutral pronouns, are syntactically diverse, and are human-validated; WinoBias includes (and delineates) sentences resolvable from syntax alone; a Winogender schema has one occupational mention and one ``other participant'' mention; WinoBias has two occupational mentions. Due to these differences, we encourage future evaluations to make use of both datasets.

\section{Conclusion and Future Work}
We have introduced ``Winogender schemas,'' a pronoun resolution task in the style of Winograd schemas that enables us to uncover gender bias in coreference resolution systems. We evaluate three publicly-available, off-the-shelf systems and find systematic gender bias in each: for many occupations, systems strongly prefer to resolve pronouns of one gender over another. We demonstrate that this preferential behavior correlates both with real-world employment statistics and the text statistics that these systems use. We posit that these systems \textit{overgeneralize} the attribute of gender, leading them to make errors that humans do not make on this evaluation.
We hope that by drawing attention to this issue, future systems will be designed in ways that mitigate gender-based overgeneralization.

It is important to underscore the limitations of Winogender schemas. As a diagnostic test of gender bias, we view the schemas as having high \textit{positive predictive value} and low \textit{negative predictive value}; that is, they may demonstrate the presence of gender bias in a system, but not prove its absence.
Here we have focused on examples of occupational gender bias, but Winogender schemas may be extended broadly to probe for other manifestations of gender bias.
Though we have used human-validated schemas to demonstrate that existing NLP systems are comparatively more prone to gender-based overgeneralization, we do not presume that matching human judgment is the ultimate objective of this line of research. Rather, human judgements, which carry their own implicit biases, serve as a lower bound for equitability in automated systems.

\section*{Acknowledgments}
The authors thank Rebecca Knowles and Chandler May for their valuable feedback on this work. This research was supported by the JHU HLTCOE, DARPA AIDA, and NSF-GRFP (1232825).  The U.S. Government is authorized to reproduce and distribute reprints for Governmental purposes. The views and conclusions contained in this publication are those of the authors and should not be interpreted as representing official policies or endorsements of DARPA or the U.S. Government.

\bibliography{naaclhlt2018}
\bibliographystyle{acl_natbib}

\clearpage

\end{document}